\documentclass{article}
\usepackage{graphicx}
\usepackage{float}
\usepackage{arxiv}
\usepackage[font=footnotesize,labelfont=bf]{caption}

\usepackage[utf8]{inputenc} 
\usepackage[T1]{fontenc}    
\usepackage{hyperref}       
\usepackage{url}            
\usepackage{booktabs}       
\usepackage{amsfonts}       
\usepackage{nicefrac}       
\usepackage{microtype}      
\usepackage{lipsum}
\usepackage[superscript]{cite}
\usepackage{bibunits}
\usepackage[labelsep=endash]{caption}
\newcounter{Figcount}
\newcounter{tempFigure}
\newenvironment{figCaption}{%
    
    \setcounter{tempFigure}{\thefigure}
    \setcounter{figure}{\theFigcount}
    }{%
    \setcounter{figure}{\thetempFigure}
    \stepcounter{Figcount}
    }

\title{\textbf{Explainable artificial intelligence model \\ to predict acute critical illness from \\ electronic health records}}

\author{
  Simon Meyer Lauritsen\textsuperscript{1,2}, Mads Kristensen\textsuperscript{1}, Mathias Vassard Olsen\textsuperscript{3}, Morten Skaarup Larsen\textsuperscript{3}, \\ \textbf{Katrine Meyer Lauritsen\textsuperscript{2},\ Marianne Johansson Jørgensen\textsuperscript{4}, 
Jeppe Lange\textsuperscript{2,4}, Bo Thiesson\textsuperscript{1,5}}
}
 
\begin{document}
\maketitle

\begin{center}

\textit{$^1$Enversion A/S, Aarhus, Denmark; $^2$Department of Clinical Medicine, Aarhus University, Denmark;\\ $^3$Department of Biomedical Engineering and Informatics, Aalborg University, Denmark;\\ $^4$Horsens Regional Hospital, Denmark; $^5$Department of Engineering, Aarhus University, Denmark.\\
*Corresponding author: Simon Meyer Lauritsen (sla@enversion.dk)  }\\
\vskip 0.2in
 
 \vskip 0.2in
\end{center}

\begin{abstract}
    We developed an explainable artificial intelligence (AI) early warning score (xAI-EWS) system for early detection of acute critical illness. While maintaining a high predictive performance, our system explains to the clinician on which relevant electronic health records (EHRs) data the prediction is grounded. Acute critical illness is often preceded by deterioration of routinely measured clinical parameters, e.g., blood pressure and heart rate. Early clinical prediction is typically based on manually calculated screening metrics that simply weigh these parameters, such as Early Warning Scores (EWS). The predictive performance of EWSs yields a tradeoff between sensitivity and specificity that can lead to negative outcomes for the patient\cite{downey2017strengths,forster2018investigating,alam2014impact}. Previous work on EHR-trained AI systems offers promising results with high levels of predictive performance in relation to the early, real-time prediction of acute critical illness\cite{tomavsev2019clinically,lundberg2019explainable,kaji2019attention,liu2019data,lauritsen2019early,barton2019evaluation,scherpf2019predicting,islam2019prediction,moor2019temporal,futoma2018learning,huang2017regular,mao2018multicentre,nemati2018interpretable,shimabukuro2017effect,kam2017learning,calvert2016computational,swaminathan2017machine,huang2017regularized}. However, without insight into the complex decisions by such system, clinical translation is hindered. In this paper, we present our xAI-EWS system, which potentiates clinical translation by accompanying a prediction with information on the EHR data explaining it.
\end{abstract}

 \vskip 0.2in

Artificial Intelligence (AI) is capable of predicting acute critical illness earlier and with greater accuracy than traditional Early Warning Score (MEWS) systems, such as modified EWSs (MEWSs) and Sequential Organ Failure Assessment Scores (SOFAs)\cite{tomavsev2019clinically,lauritsen2019early,barton2019evaluation,islam2019prediction,moor2019temporal,futoma2018learning,mao2018multicentre,shimabukuro2017effect,kam2017learning,calvert2016computational,futoma2017improved,futoma2017learning,vellido2017machine,ten2017risk,shickel2019deepsofa}. Unfortunately, standard deep learning (DL) that comprise available AI models are black-boxes and their predictions cannot readily be explained to clinicians. The importance of explainable and transparent DL algorithms in clinical medicine is without question and was recently highlighted in the Nature Medicine review by Topol, E. J.\cite{topol2019high,shickel2017deep} 
Transparency and explainability are an absolute necessity for the widespread introduction of AI models into clinical practice, where an incorrect prediction can have grave consequences\cite{shickel2017deep,rajkomar2019machine,xiao2018opportunities,cabitza2017unintended}. Clinicians must be able to understand the underlying reasoning of AI models so they can trust the predictions and be able to identify individual cases in which an AI model potentially gives incorrect predictions\cite{shickel2017deep,rajkomar2019machine,xiao2018opportunities,the2018opening}.

\begin{figure}[!h]
  \centering
  \includegraphics[width=\textwidth]{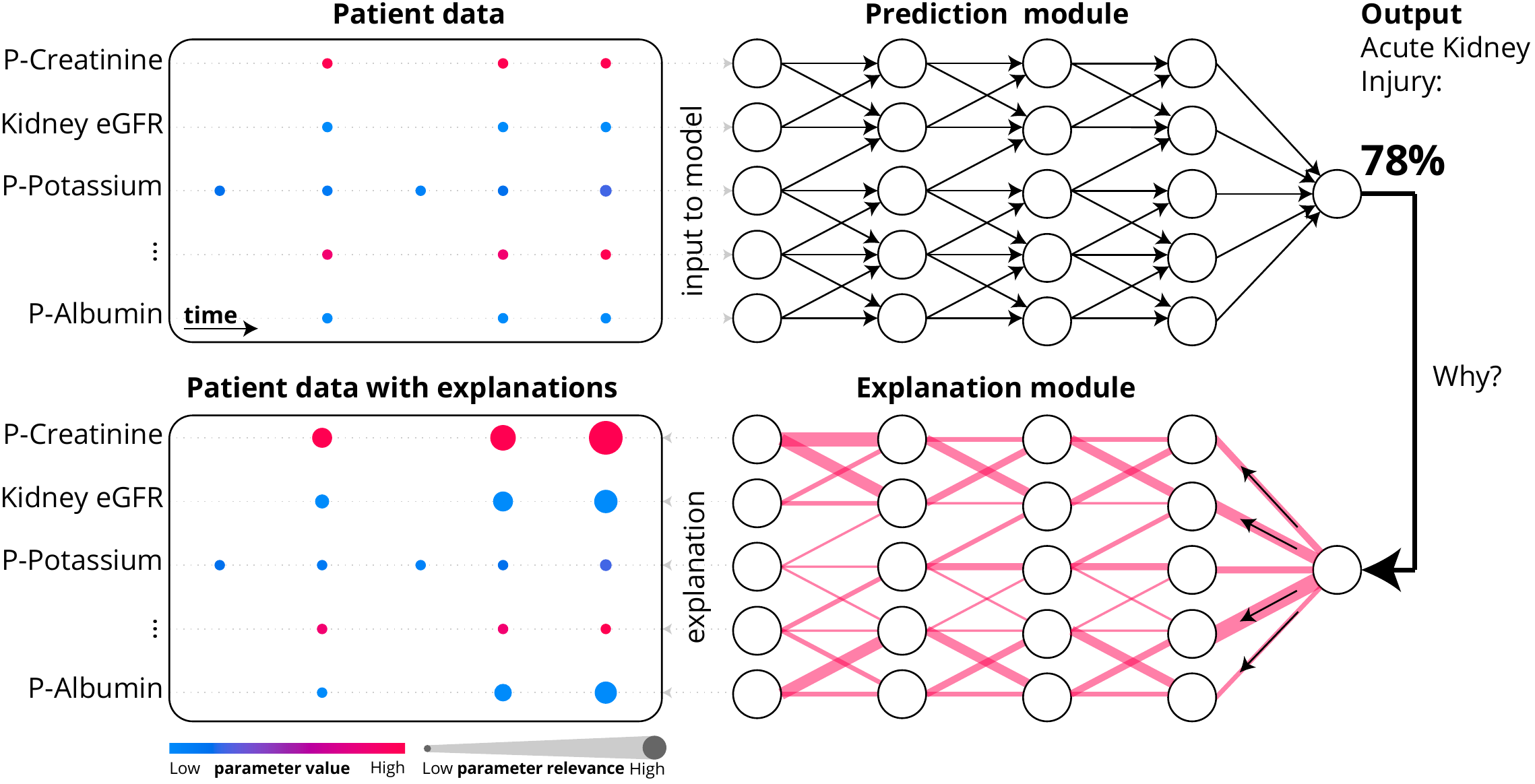}
  \caption{\textbf{Overview of the xAI-EWS system.} Each patient’s data from the EHR is used as input in the TCN prediction module. Based on this data, the model makes a prediction, such as a 78\% risk of AKI. The DTD explanation module then explains the TCN predictions in terms of input variables. Plasma: P; estimated Glomerular filtration rate: eGFR; Deep Taylor Decomposition: DTD; temporal convolutional network: TCN, and the Explainable Artificial Intelligence Early Warning System: xAI-EWS}
  \label{fig:fig1}
\end{figure}

In this paper, we will present xAI-EWS, which comprises a robust and accurate AI model for predicting acute critical illness from Electronic Health Records (EHRs). Importantly, xAI-EWS was designed to provide explanations for the given predictions. To demonstrate the general clinical relevance of the xAI-EWS, we present results here from three emergency medicine cases: sepsis, Acute Kidney Injury (AKI), and Acute Lung Injury (ALI). The xAI-EWS is composed of a temporal convolutional network (TCN)\cite{kalchbrenner2016neural,bai2018empirical,lea2016temporal} prediction module and a deep Taylor decomposition (DTD)\cite{montavon2017explaining,montavon2018methods,bach2015pixel,samek2017explainable,samek2016evaluating} explanation module, tailored to temporal explanations (see figure \ref{fig:fig1}).
The architecture of the TCN has proven to be particularly effective at predicting events that have a temporal component, such as the development of critical illness\cite{moor2019temporal,bai2018empirical,lea2016temporal,aksan2019stcn,pedersen2011danish}. The TCN operates sequentially over individual EHRs and outputs predictions in the range of 0\% to 100\%, where the predicted risk should be higher for those patients at risk of later acute critical illness, compared to those who are not.\\
The DTD explanation module delineates the TCN predictions in terms of input variables by producing a decomposition of the TCN output on the input variables. As comparative measures for the predictive performance, we used the area under the receiver operating characteristic curve (AUROC) and the area under the precision-recall curve (AUPRC). Regarding the explanations, the quality was assessed by manual inspection of trained specialists (medical doctors) in emergency medicine. The xAI-EWS system was trained using data from a multi-center Danish data set, which included the health information of citizens aged 18 years and older with residency in one of four Danish municipalities for the period of 2012 to 2017. The dataset contained information from EHRs related to biochemistry, medicine and microbiology from a diverse set of clinical departments. A total of 163,050 admissions (45.86\% male), distributed across 66,288 unique citizens were included for analysis. Patients were randomized across training (80\%), validation (10\%), and test (10\%) sets. The xAI-EWS was validated using a five-fold cross-validation. A ground-truth label for the presence of sepsis, AKI, and ALI at any given point in time was added using the criteria of the Third International Consensus Definitions for Sepsis (Sepsis-3)\cite{singer2016third}, the Kidney Disease: Improving Global Outcomes (KDIGO)\cite{khwaja2012kdigo} , and the need for noninvasive ventilation (NIV) or continuous positive airway pressure (CPAP), respectively. Among the admissions, the prevalence for sepsis, AKI, and ALI were 2.44\%, 0.75\%, and 1.68\%, respectively. We implemented two clinical baseline models (MEWS and SOFA) and a gradient-boosting model.

For each patient, data from 34 chosen clinical parameters were extracted from the EHRs and used as input for the prediction module (Extended Data Table 1). Based on this data, each of the three prediction models output a prediction in the range of 0\% to 100\%, such as a 78\% risk of developing AKI during admission. The DTD explanation module was then used to explain individual patient predictions in terms of the clinical parameters measured during their admission. The module produced an explanation by running a backward pass on the prediction output according to a predefined set of propagation rules (Figure 1).
\begin{figure}[!h]
  \centering
  \includegraphics[width=\textwidth]{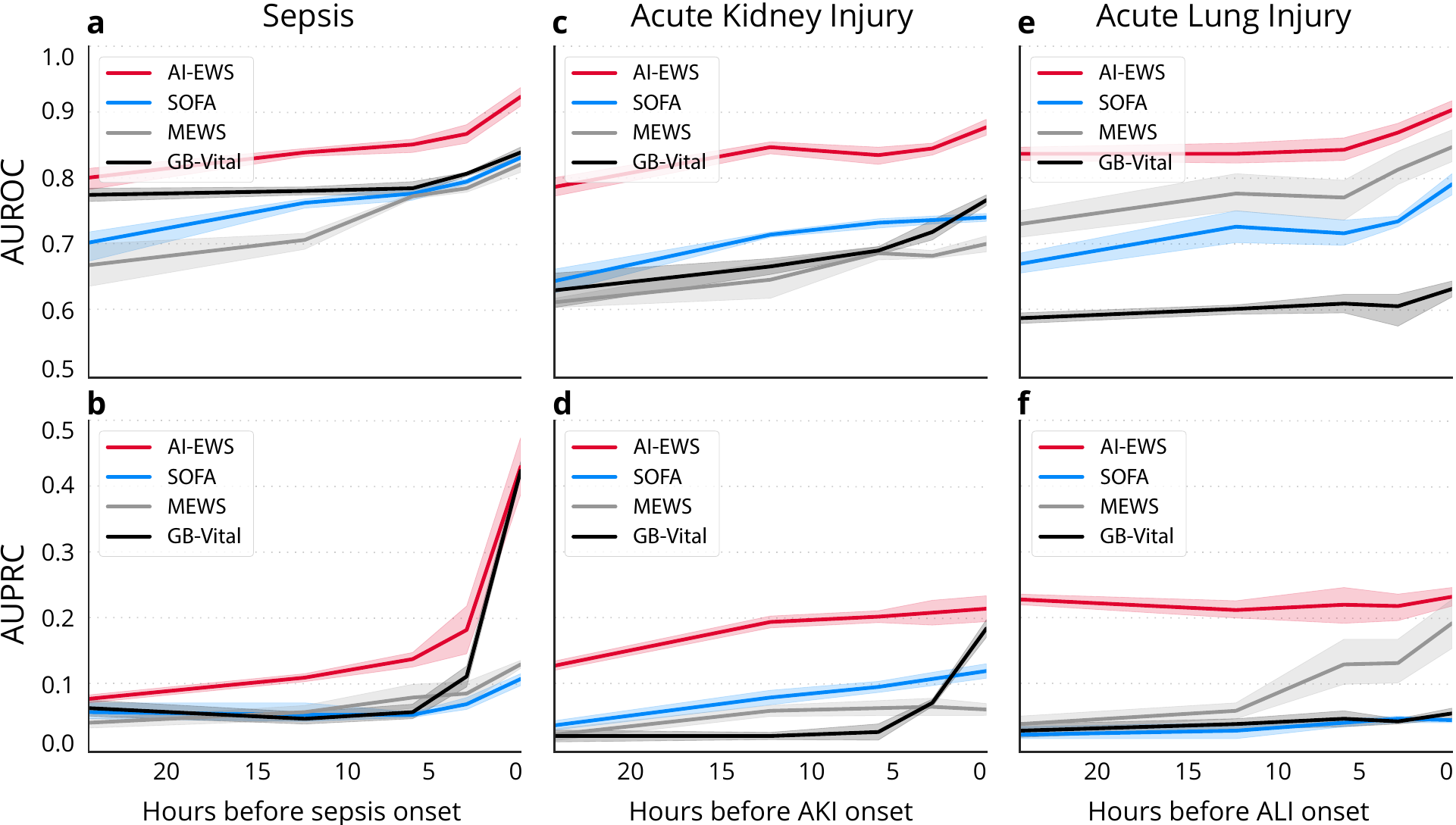}
  \caption{\textbf{Predictive performance of the xAI-EWS.} The xAI-EWS results are compared with those from MEWS, SOFA, and the Gradient Boosting Vital Sign Model (GB-Vital). Predictive performance is shown from the onset time to 24 hours before onset. AUROC performance is shown for sepsis (\textbf{a}), AKI (\textbf{c}), and ALI (\textbf{e}), and AUPRC performance is shown for sepsis (\textbf{b}), AKI (\textbf{d}), and ALI (\textbf{f}). Lighter colors surrounding the lines indicate uncertainty by 95\% confidence intervals calculated from the five-fold cross-validation.}
  \label{fig:fig2}
\end{figure}

In Figure 2, the predictive power of the xAI-EWS is presented in summary form with results from the onset time to 24 hours before onset. AUROC with mean values and 95\% Confidence Intervals (CIs) were 0.92(0.9-0.95)–0.8(0.78-83), 0.88(0.86–0.9)–0.79(0.78–0.8), and 0.90(0.89–0.92)–0.84(0.82–0.85) for sepsis, AKI, and ALI, respectively. AUPRC with mean values and 95\% CIs were 0.43(0.36–0.51)–0.08(0.07-0.09), 0.22(0.19-0.24)-0.13(0.12-0.14), and 0.23(0.21-0.26)–0.23(0.22-0.24) for sepsis, AKI, and ALI, respectively. (Extended Data Table 3 and 4).
The xAI-EWS enabled two perspectives on the model explanations: an individual and a population-based perspective. For the individual perspective, the explanation module enabled the xAI-EWS to pinpoint which clinical parameters at a given point in time were relevant for a given prediction. In current clinical practice, the workflow normally follows that clinicians observe either a high EWS or an increase in EWS. However, the following targeted clinical intervention concerning the potential critical illness happens when the clinician understands which clinical parameters have caused the high EWS or the change in EWS. This is one of the main reasons why AI-based EWS systems need to be able to explain their predictions. The xAI-EWS system we developed allows for such explanations in real time and across all clinical parameters used in the model. An example of an output from the explanation module, utilizing the individual perspective, is illustrated in Figure 3. Individual clinical parameters are sized according to the amount of backpropagated relevance. Figure 3\textbf{a} shows the 10 most relevant parameters with respect to sepsis for a single patient with a risk score of 76.2\%. High respiration frequency, high pulse rate, and low plasma albumin are the most important predictors of sepsis. The physiological values of respiration frequency and pulse rate do not seem to increase close to the prediction time, but, inspecting the increasing sizes of relevance, it appears that the xAI-EWS attributes more weight to recent values. Figure 3\textbf{b} and Figure 3\textbf{c} show the 10 most relevant parameters with respect to AKI and ALI for two patients with risk scores of 90.4\% and 83.8\%, respectively.

\begin{figure}[!h]
  \centering
  \includegraphics[width=\textwidth]{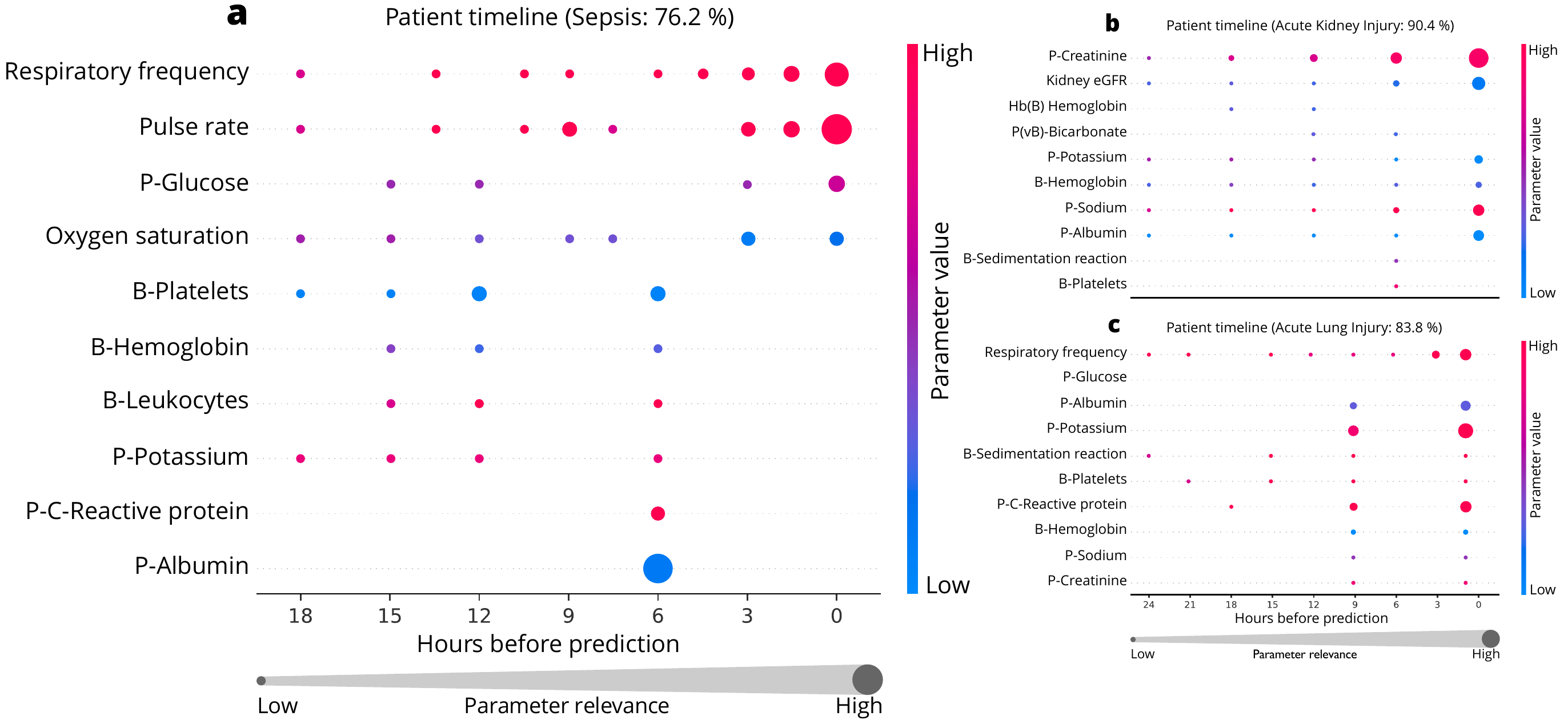}
  \caption{\textbf{Results from the explanation module displays for three individual patients.} Three selected patient timelines with back-propagated relevance for sepsis (\textbf{a}), AKI (\textbf{b}), and ALI (\textbf{c}) are shown. Only the 10 highest-ranking parameters in descending order by the mean relevance are displayed. The data shown in the three timelines match the data from the observation window, such that a time equal to zero is the prediction time. The data-points are colored according to the 5th and 95th percentiles for each parameter across the whole dataset. The blue data-points correspond to a value between 0 and the 5th percentile, the red data-points correspond to a value between the 95th and 100th percentiles, and data-points with values close to the median are purple. Plasma: P; estimated Glomerular filtration rate: eGFR.}
  \label{fig:fig3}
\end{figure}

In terms of the population-based perspective, the xAI-EWS is able to facilitate transparency and, thereby, induce trust, by giving clinicians insights into the internal mechanics of the model without any deep technical knowledge of the mechanisms behind it. 
In Figure 4, the 10 most important clinical parameters for each of the three models are shown. The parameters are sorted by the decreasing mean relevance as computed for the local, back-propagated relevance scores across the entire population, but only for patients who were positive for sepsis, AKI, or ALI.

The blue horizontal bars in the left column of Figure 4 display the mean relevance. In the local explanation summary in the right-hand column of Figure 4, the distribution of the back-propagated relevance scores for each clinical parameter are shown and color-coded by the parameter value associated with the local explanation. As an example, in Figure 4\textbf{c} and Figure 4\textbf{d}, the AKI model seems to associate high P-creatinine levels and low estimated glomerular filtration rates with AKI. When the model is confident about a decision, it will output a high probability. This high probability will result in more relevance available for distributing backward; it will also result in larger relevance scores. On the contrary, when the model does not believe that a patient will develop an acute critical illness, it will output a low probability, and the associated relevance scores will also be low. The summary distribution allows clinicians to get an idea of what to expect from the model in clinical practice.
It is important to note that the xAI-EWS presented in this study should not be conceived as the one-and-only multi-outcome model. Rather, it should be viewed as a general method of building precise and explainable models for acute critical illness. Following this line of thinking, it is obvious that other models with important critical outcomes, such as hypokalemia, hyperkalemia, acute constipation, and cardiac arrest, should be added to the three models presented in this study. This will result in a series of EWS models that are all specialists in their respective fields.

One important aspect on the basis of our model is that more work is needed to investigate better ground truth definitions of the evaluated critical illnesses, such as AKI and ALI. We based the ground truth on the need for Continuous Positive Airway Pressure (CPAP) or Noninvasive ventilation (NIV) due to the lack of a convention on how to label ALI. The KDIGO is an indicator of AKI that has a long lag time after the initial renal impairment, as mentioned by Tomašev et al.\cite{tomavsev2019clinically} Our model is trained and has been tested on a large dataset that is highly representative of the Danish population. However, validating the predictive performance of the xAI-EWS on a different population would make for an interesting study, and, as the xAI-EWS currently uses just 34 clinical parameters, this appears feasible.

In summary, we have presented the xAI-EWS—an explainable AI EWS system for the prediction of acute critical illness using EHRs. The xAI-EWS shows a high predictive performance while enabling the possibility to explain the predictions in terms of pinpointing decisive input data to empower clinicians to understand the underlying reasoning of the predictions. We hope that our results will be a steppingstone toward a more widespread adoption of AI in clinical practice. As stated, explainable predictions facilitate trust and transparency—properties that also make it possible to comply with the regulations of the European Union General Data Protection Regulation, the Conformité Européenne (CE) marking, and the United States Food and Drug Administration (FDA)\cite{kaminski2019right,selbst2017meaningful}.

\begin{figure}[!htbp]
  \centering
  \includegraphics[width=6in]{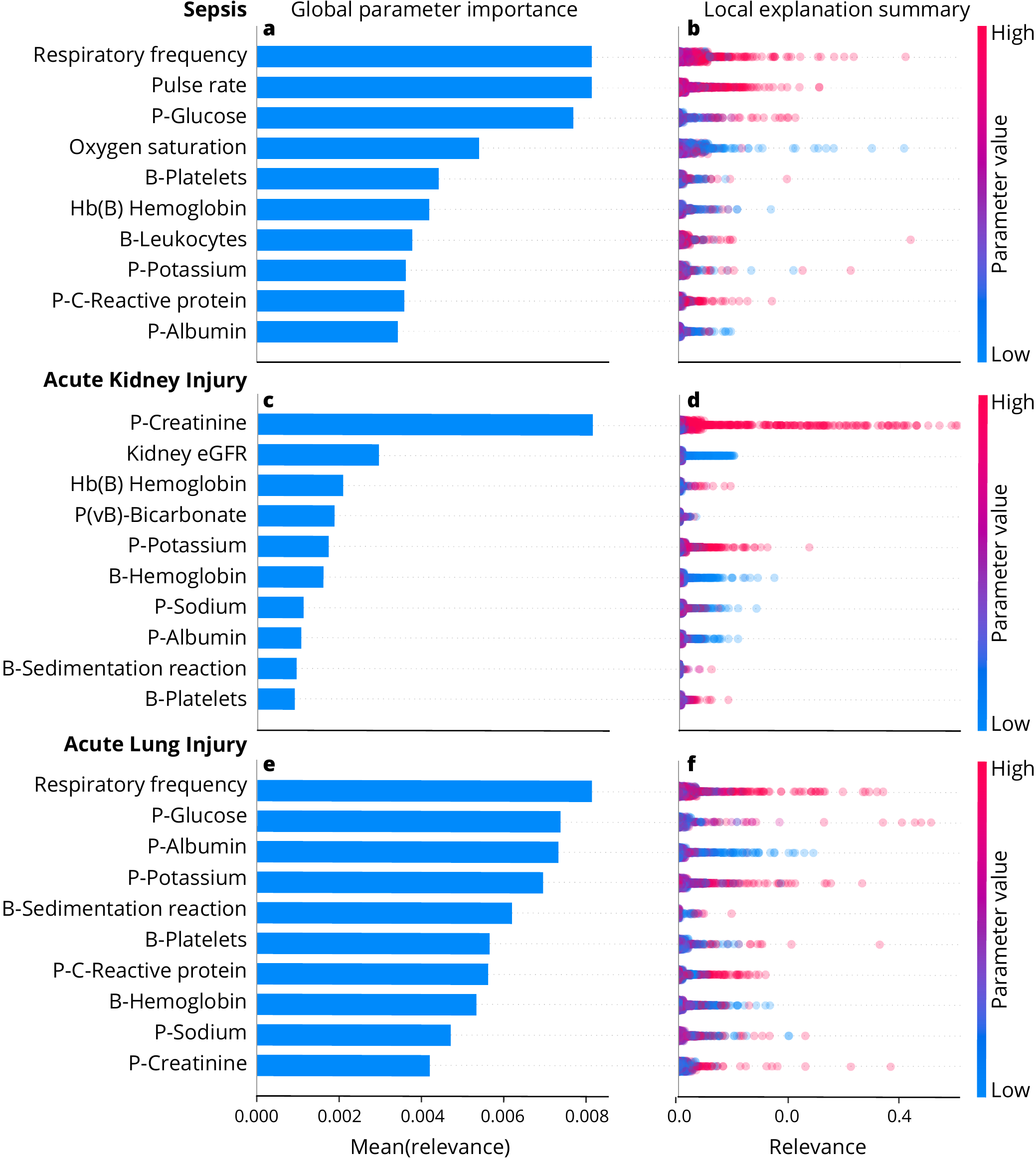}
  \caption{\textbf{Results from the explanation module displaying the global parameter importance and local explanation summary.} The parameters are sorted in descending order according to global parameter importance, defined by the mean relevance, which is identified by the blue horizontal bars for sepsis (\textbf{a}), AKI (\textbf{c}), and ALI (\textbf{e}). The local explanations summary shows all the individual data-points, colored by parameter value and displaced by the relevance for sepsis (\textbf{b}), AKI (\textbf{d}), and ALI (\textbf{f}). The height of the data-points shown for each parameter in the local explanation summary correlate with the number of data-points at the associated level of relevance. The population-based perspective is simplified by ignoring the temporal relevance variations, treating all data-points at different times equally. Plasma: P; estimated Glomerular filtration rate: eGFR.}
  \label{fig:fig4}
\end{figure}








\newpage
{\LARGE Methods}

\textbf{Data description}\\
In this study, we analyzed the secondary healthcare data of all residents of four Danish municipalities (Odder, Hedensted, Skanderborg, and Horsens) who were 18 years of age or older for the period of 2012–2017 \cite{cross}. The data contained information from the electronic health record (EHR), including biochemistry, medicine, microbiology, and procedure codes, and was extracted from the “CROSS-TRACKS” cohort, which embraces a mixed rural and urban multi-center population with four regional hospitals and one larger university hospital. Each hospital comprises multiple departmental units, such as emergency medicine, intensive care, and thoracic surgery. We included all 163,050 available inpatient admissions (45.86\% male) during the study period and excluded only outpatient admissions. The included admissions were distributed across 66,288 unique residents. The prevalence for sepsis, AKI, and ALI among these admissions was 2.44\%, 0.75\%, and 1.68\%, respectively (see Extended Data Table 2). The CROSS-TRACKS cohort receives data from the regional data warehouse (Central Denmark Region), which offers a combined dimensional model of the secondary healthcare data. Merging all data sets is possible via a unique personal identification number given to all Danish citizens and by which all information within any public institution is collected\cite{pedersen2011danish}.
The model parameters were limited to include only a small subset of the total available events, namely, 28 laboratory parameters and six vital signs (see Extended Data Table 1). The parameters were selected by trained specialists in emergency medicine (medical doctors) with the sole purpose of simplifying the model to enable a better discussion of the model explanations. Therefore, frequently recorded parameters having known correlations with acute critical illnesses were preferred. While a deeper model with a latent representation of thousands of clinical parameters might lead to better performance, it would also have made the continuous discussions between clinicians and software engineers difficult. Therefore, the scope of this article is not to obtain the best performance at all costs but to demonstrate how clinical tasks can be supported by a fully explainable deep learning approach. 

\textbf{Data preprocessing.} In the data extracted from the CROSS-TRACKS cohort, each admission is represented as a time-ordered sequence of EHR events. Importantly, the time-stamped order of this data reflects the point in time at which the clinicians record each event during the admission. Each event comprises three elements: a time stamp; an event name, such as blood pressure; and a numerical value. The event sequence is partitioned in aggregated intervals of one hour, such that the observation window of 24 hours is divided into 24 one-hour periods, and all the events occurring within the same one-hour period are grouped together by their average numerical value.  

\textbf{Gold standards.} Via a classification process, each admission was classified as sepsis-positive, AKI-positive, ALI-positive, or negative (no critical illness). For sepsis classification, we followed the recent Sepsis-3\cite{singer2016third,seymour2016assessment} implementation by Moor et al.\cite{moor2019temporal}, according to which both suspected infection and organ dysfunction are required to be present\cite{moor2019temporal,singer2016third,seymour2016assessment}. Suspected infection was defined by the co-occurrence of body fluid sampling and antibiotic administration. When a culture sample was obtained before antibiotics administration, the antibiotic had to be ordered within 72 hours\cite{moor2019temporal}. If the antibiotic was administered first, then the culture sample had to follow within 24 hours\cite{seymour2016assessment}. Organ dysfunction occurs when the SOFA\cite{vincent1996sofa} score displays an acute increase of more than or equal to two points\cite{seymour2016assessment}. To implement the organ dysfunction criterion, we used a 72-hour window from 48 hours before to 24 hours after the time of suspected infection, as suggested by Singeret et al.\cite{singer2016third} and Moor et al.\cite{moor2019temporal}. The Sepsis-3 implementation is visualized in Extended Data Figure 1.

\begin{figure}[!h]
\begin{figCaption}
  \centering
  \includegraphics[width=\textwidth]{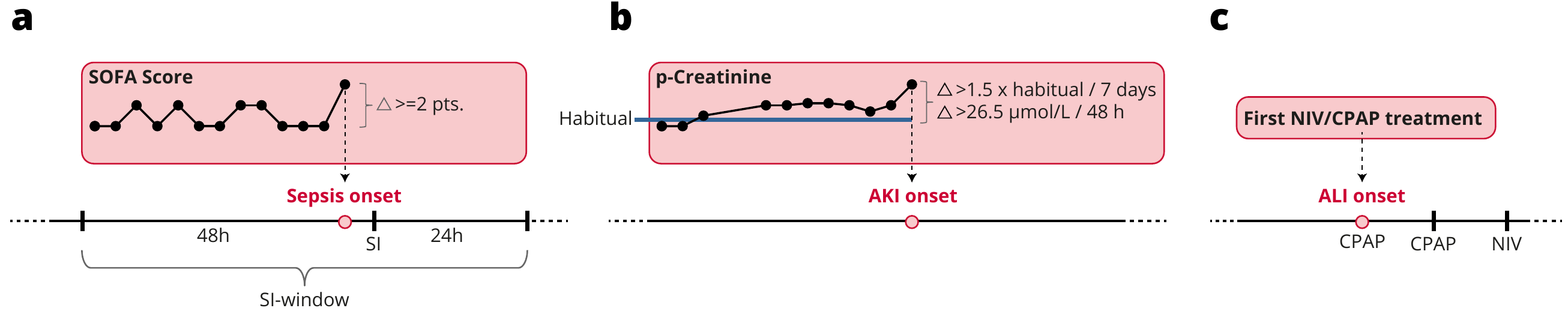}
  \caption{\textbf{Gold standards for sepsis, AKI, and ALI.} Sepsis (\textbf{a}); AKI (\textbf{c}); ALI (\textbf{e}); Suspected infection (SI).}
  \label{fig:fig5}
  \end{figCaption}
\end{figure}

AKI classification was performed according to the KDIGO criteria\cite{khwaja2012kdigo}. KDIGO accepts three definitions of AKI: (1) an increase in serum creatinine of 0.3 mg/dl $(26.5 \mu mol/l)$ within 48 hours; (2) an increase in serum creatinine by 1.5 times the habitual creatinine level of a patient within the previous seven days; and (3) a urine output of < 0.5 ml/kg/h over 6 hours. Following the work of Tomašev et al.\cite{tomavsev2019clinically}, only the first two definitions were used to provide ground-truth labels for the onset of AKI as urine measurements were not available. The habitual creatinine level was computed as the mean creatinine level during the previous 365 days. We used binary encoding for AKI such that all three severity stages (KDIGO stages 1, 2, and 3) were encoded as positive AKI. For ALI classification, we simply considered the presence of either NIV or CPAP during the admission. The ALI onset was the first occurrence of either NIV or CPAP (see Extended Data Figure 1).

\textbf{Prediction module.} The AI-EWS model is designed as a variation of a convolutional neural network (CNN) called a temporal convolutional network (TCN). CNNs have dominated computer vision tasks for the last century\cite{the2018opening} and are also highly capable of performing sequential tasks, such as text analysis and machine translation\cite{bai2018empirical,zhang2015character}. A TCN\cite{bai2018empirical,lea2016temporal} models the joint probability distribution over sequences by decomposing the distribution over discrete time-steps $p_\theta(\textbf{x})=\amalg_{t=1}^Tp_\theta (x_t|x_{1:t-1})$, where $\textbf{x}=\textbf{\{}x_1,x_2,…,x_T\textbf{\}}$ is a sequence, and the joint distribution is parameterized by the TCN parameter $\theta$. Thus, a TCN operates under the autoregressive premise that only past values affect the current or future values, e.g., if a patient will develop acute critical illness. Moreover, TCNs differ from “ordinary” CNNs by at least two properties: (1) the convolutions in TCNs are causal in the sense that a convolution filter at time t is only dependent on the inputs that are no later than t, wherein the input subsequence is$x_1,x_2,…,x_t$; and (2) TCNs can take a sequence of any length as input and output a sequence of the same length, similar to RNNs \cite{lea2016temporal}. The TCN achieves this by increasing the receptive field of the model with dilated convolutions instead of performing the traditional max pooling operation, as seen in most CNNs. Dilated convolutions achieve a larger receptive field with fewer parameters by having an exponential stride compared to the traditional linear stride. By increasing the receptive field, a temporal hierarchy comparable to multi-scale analysis from computer vision can be achieved\cite{crowley1987multiple}. The Extended Data in Figure 2 schematizes the xAI-EWS model and the concept of dilated convolutions. At the time of prediction, the xAI-EWS model receives an input matrix of shape "time-steps" $\times$ "features"  for each patient.

\begin{figure}[!h]
\begin{figCaption}
  \centering
  \includegraphics[width=\textwidth]{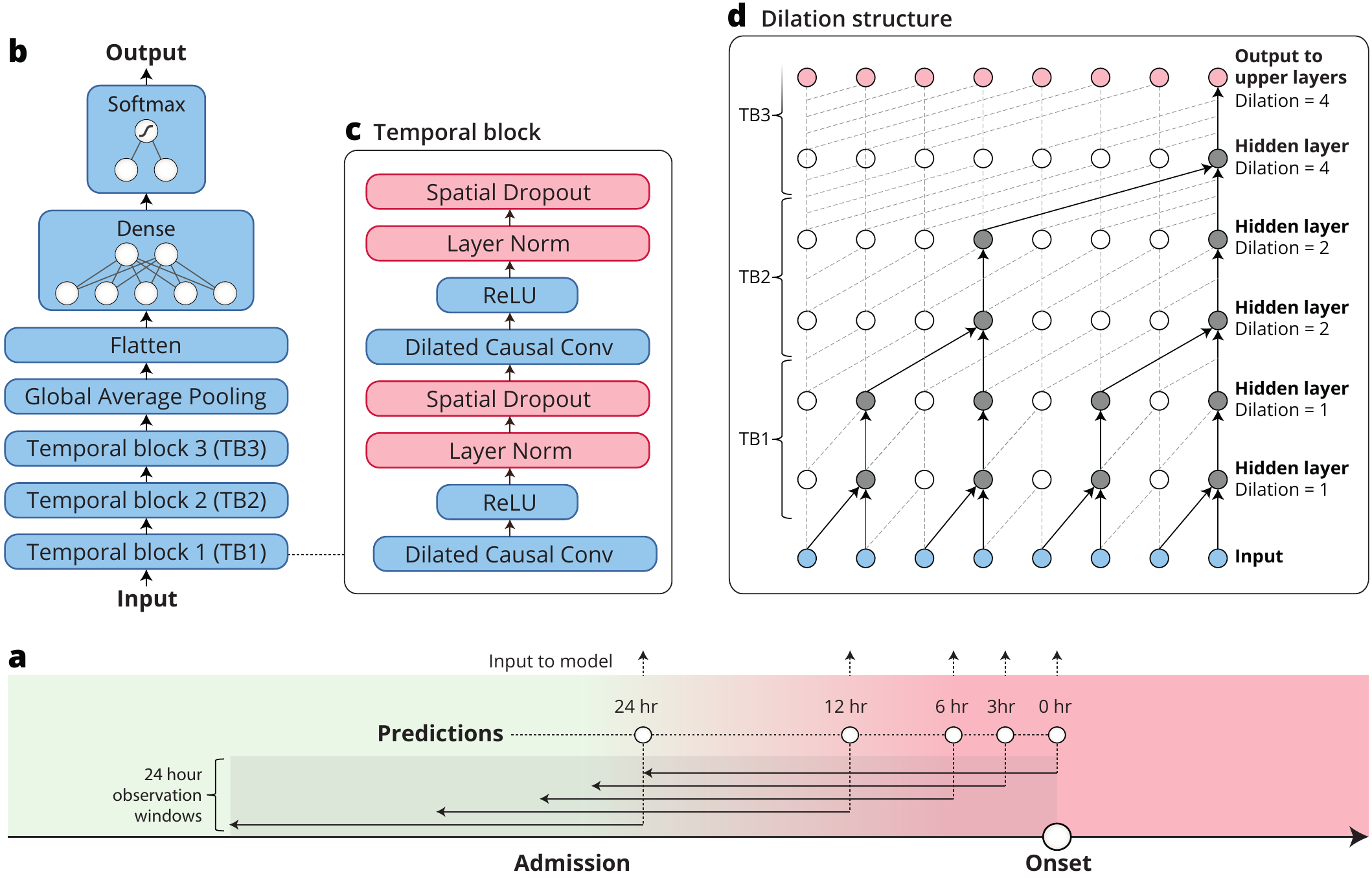}
  \caption{\textbf{The xAI-EWS model architecture.} The models in this study are trained and evaluated at 0, 3, 6, 12, and 24 hours before the onset of critical illness. Each model has a 24-hour retrospective observation window. The color gradient from green to red illustrates continuous deterioration towards acute critical illness (\textbf{a}). The overall model architecture of the AI-EWS model is shown in (\textbf{b}). The xAI-EWS uses three temporal blocks (\textbf{c}), each comprising one-dimensional dilated causal convolution layers, ReLU activations, one-dimensional dropout layers, and normalization layers. Red layers are only used during training and are omitted when the model is used for predictions and explanations. The overall dilation structure of the model is shown in (\textbf{d}). The one-dimensional dilated causal convolution layers allow the model to skip some points during convolution and, thereby, increase the receptive field of the model.}
  \label{fig:fig6}
  \end{figCaption}
\end{figure}

The data is processed by three temporal blocks, each including one-dimensional dilated causal convolutions with 64 filters, ReLU activations\cite{nair2010rectified}, layers normalization\cite{ba2016layer}, and one-dimensional spatial dropout layers\cite{tompson2015efficient}. Outputs from the third temporal block are pooled together across time-steps by a global average pooling operation\cite{lin2013network} to obtain a stabilizing effect for the final output of the model. The pooled output from each kernel in the dilated causal convolutions is flattened to a single vector that is used as input to a final dense layer followed by a softmax activation function. The output from the softmax activation is the probability of future sepsis, AKI, or ALI during admission.

\textbf{Training and hyperparameters.} The model was trained to optimize the cross-entropy loss using the Adam optimizer\cite{kingma2014adam} with mini-batches of the size of 200, a learning rate of 0.001, and a dropout rate of 10\%. All weights were initialized with He Normal initialization\cite{he2016deep}. The model was trained on a NVIDIA Tesla V100 GPU. Convergence was reached in approximately 30 minutes.

\textbf{Explanation module}. In linear models, such as linear regression models33 and support vector machines (with a linear kernel)\cite{patle2013svm}, the association between parameters and outcomes is modeled linearly and is, therefore, readily transparent and explainable. Consider the linear function $f_c$ that weights the input \textbf{x} by $\textbf{w}_c$ in order to assign a decision for class c, where, for notational convenience, the bias is included in the input vector with a fixed weight of one:

\begin{equation}
f_c(\textbf{x})=\textbf{w}_c^T\textbf{x}=\sum_iw_{ic}x_i.    
\end{equation}

From this expression, we can conclude that each input $x_i$ of \textbf{x} contributes together with the trainable parameter $w_{ic}$ to the overall evaluation of $f_c$ through the quantity $w_{ic}x_i$. The weight $w_{ic}$ can therefore be interpreted as a factor that reflects the importance of the parameter $x_i$, and this interpretation, thereby, offers a simple explanation for each decision made by the linear model.  In contrast, the complexity associated with the multi-layer non-linear nature of deep learning models counteracts with such simplicity in explanations.
Layer-wise relevance propagation (LRP)\cite{montavon2017explaining,montavon2018methods,samek2017explainable,ba2016layer,tompson2015efficient,lin2013network} is an explantory technique that applies to deep-learning models, such as TCNs. Starting from the output $f_c(\textbf{x})$, LRP decomposes an explanation into simpler local updates, each recursively defining the contribution to the explanation (called relevance) for all activating neurons in the previous layer. The relevance explanation of a prediction is, in this way, propagated backward from the initial relevance score $R_j=f_c(\textbf{x})$ through the network by local relevance updates $R_{i\leftarrow j}$ between connecting neurons i and j until the input layer is finally reached. Extended Data Figure 3 illustrates this process, which is similar to standard back-propagation of errors except that the relevance data are propagated backward in the network instead. 
The so-called deep Taylor decomposition (DTD)\cite{lea2016temporal} defines a sound theoretical framework behind most implementations of LRP. In DTD, a local backward propagation of relevance from a neuron to the activating neurons in the previous layer resembles the above explanation for the simple linear model except that the DTD relevance-propagation rule accounts for non-linearity in the model by a first-order Taylor approximation at some well-chosen root-point. For models where linear projections are followed by non-linear ReLU activation functions, such as this work, the standard update rule for propagating the relevance $R_j$ at neuron j backwards to neuron i is given by
\begin{equation}
	R_{i\leftarrow j}=\frac{w_{ij}^+a_i}{\sum_iw_{ij}^+a_i}R_j,   
\end{equation}

where $a_i$ is the activation for neuron i, and $w_{ij}^+=w_{ij}$ for positive weights and otherwise equals zero. Notice that, aside from the restriction to positive weights, this local relevance update rule is similar to the simple explanation rule for the linear model.

\begin{figure}[!h]
\begin{figCaption}
  \centering
  \includegraphics[width=5in]{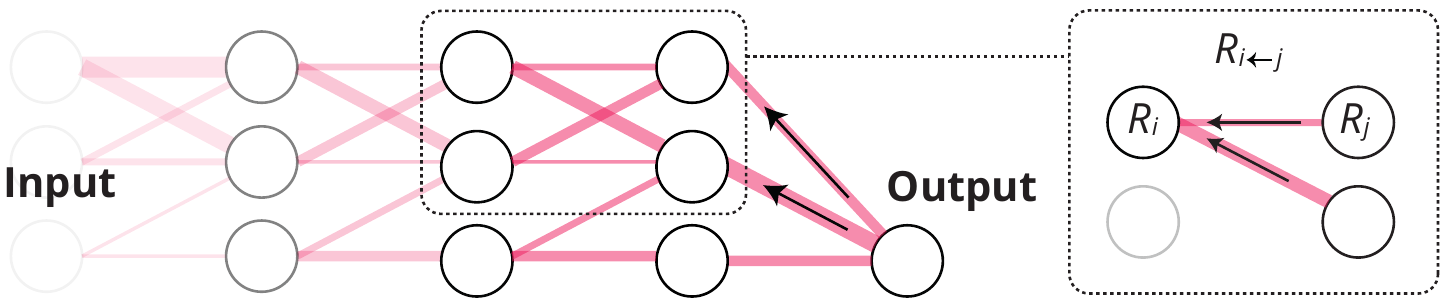}
  \caption{\textbf{Deep Taylor Decomposition (DTD).} DTD decomposes the problem of explaining a complex multilayer neural network model into simpler sub-functions that are easier to analyze and explain. The relevance score at input neuron $R_i$ is obtained by pooling all incoming relevance values $R_j$ from the output neurons in the next layer.}
  \label{fig:fig7}
  \end{figCaption}
\end{figure}

Finally, to propagate the relevance at neuron i further back in the network, the collective relevance score $R_i$  is obtained by pooling all incoming relevance values $R_{i\leftarrow j}$ from the output neurons to which i contributes in the forward pass.
\begin{equation}
    R_i=\sum_jR_{i\leftarrow j}
\end{equation}

Together, Equations (2) and (3) define the DTD procedure, where the relevance propagation rule in (2) is a special instance of DTD. In general, the Taylor expansion defining the relevance propagation rule depends on the type of non-linearity and can, in addition, be engineered to enforce desirable properties based on root point restrictions. For more details about DTD, please refer to the original work by Montavon et al.\cite{lea2016temporal} or the collection of works in Montavon et al.\cite{he2016deep}

The AI-EWS explanation module allows two perspectives on model explanations: an individual and a population-based perspective. For the individual perspective, DTD can be used for all patients with a high probability of developing acute critical illness. The module will simply pinpoint which clinical parameters at a given point in time were relevant for any given prediction (Figure 3). For the population-based perspective, relevance is back-propagated from the output neuron representing the positive classes (sepsis, AKI, and ALI) and is only considered for the patients with a positive label (sepsis, AKI, and ALI). The individual data points and back-propagated relevance scores for these patients were aggregated in two ways to enable global parameter importance estimation and local explanation summary\cite{lundberg2019explainable} (\ref{fig:fig4}). For estimating global parameter importance, the mean relevance scores associated with each clinical parameter were computed. This computation enabled parameter-importance estimation comparable to standardized regression coefficients in multiple linear regression or feature importance measures in random forest\cite{montavon2017explaining}. The local explanation summary (Figure 4\textbf{b}) presents all individual data points, colored by parameter value and displaced by the relevance. In the local explanation summary, the height of the data points shown for each parameter correlates with the number of data points at their associated level of relevance. The population-based perspective is simplified by ignoring potential temporal relevance variations and treating all data points at different times equally. The visual concepts of global parameter importance estimation and local explanation summary used in this paper are adopted from the SHapley Additive exPlanations (SHAP)\cite{lundberg2019explainable} library by Lundberg et al. In this paper, DTD was implemented using the iNNvestigate\cite{alber2019innvestigate} library developed by Alber et al. iNNvestigate is a high-level library with an easy-to-use interface for many of the most-used explanation methods for neural networks.

\textbf{Baseline models.} MEWS. The MEWS baseline model interprets raw MEWS scores as a prediction model for acute critical illness. MEWS was implemented as the Danish variant called “Early detection of critical illness” [TOKS: Tidlig opsporing af kritisk sygdom]\cite{perspektiv2018gennemgang}. MEWS scores were calculated each time one of the model components was updated with a new measurement. Missing values were imputed with a standard carry-forward interpolation. 

SOFA. This baseline model interprets raw SOFA\cite{singer2016third,seymour2016assessment} scores as a prediction model for acute critical illness. SOFA scores were calculated each time one of the model components was updated with a new measurement. Missing values were imputed with a standard carry-forward interpolation.

GB-Vital. This baseline model is a replication of a well-known sepsis detection mode\cite{calvert2016computational,mao2018multicentre} from the literature, which has shown great results in a randomized study\cite{shimabukuro2017effect}. The complete technical description of the model can be found in the study from Mao et al\cite{mao2018multicentre}. The model parameters are constructed by considering only six vital-sign events from the raw EHR event sequences: systolic blood pressure, diastolic blood pressure, heart rate, respiratory rate, peripheral capillary oxygen saturation, and temperature. The constructed parameters highly aggregate the sequence information, and only limited ordering information is retained. That is, for each of the six vital signs, five parameters are constructed to represent the average value for the current hour, the prior hour, and the hour prior to that hour, together with the trend value between two succeeding hours. Based on these 30 parameters (five parameters from each of the six vital-sign events), the GB-Vital model is constructed as a gradient-boosted classifier of decision trees61. 

\textbf{Evaluation.} The xAI-EWS model outputs a prediction that reflects the risk of developing critical illness during a hospital admission if not intervened upon. This prediction ranges from zero to one, where the predicted risk should be higher for those patients who are at a risk of developing a critical illness in the future than the patients who are not at risk. The xAI-EWS model was validated using five-fold cross-validation. For each fold, the data were divided into training data (80\%), validation data (10\%), and test data (10\%). The training data were used to fit the xAI-EWS model parameters. The validation data were used to perform an unbiased evaluation of a model fit during training, and the test data were used to provide an unbiased evaluation of the final model. As comparative measures, we used the area under the receiver operating characteristic curve (AUROC) and the area under the precision-recall curve (AUPRC). 

\textbf{Ethics and information governance.} The study was approved by The Danish Data Protection Agency [case number 1-16-02-541-15]. Additionally, the data used in this work were collected with the approval of the steering committee for CROSS-TRACKS. Only retrospective data were used for this research without the active involvement of patients or potential influence on their treatment. Therefore, under the current national legislature, no formal ethical approval was necessary.
  
\textbf{Data availability}\\
The authors have accessed the data referred to herein by applying the CROSS-TRACKS cohort, which is a newer Danish cohort that combines primary and secondary sector data. Due to the EU regulations, GDPR, these data are not readily available to the wider research community per se. However, all researchers can apply for access to the data by following the instructions on this page:\url{ http://www.tvaerspor.dk/adgang-til-data}.

\textbf{Code availability}\\
We made use of several open-source libraries to conduct our experiments: the high-level library explaining neural networks iNNvestigate (\url{https://github.com/albermax/innvestigate}); the machine learning framework TensorFlow (\url{https://github.com/tensorflow/tensor-flow}); and SHAP (\url{https://github.com/slundberg/shap}). Our experimental framework makes use of proprietary libraries, and we are unable to publicly release this code. We have described the experiments and implementation details in the Methods section to allow for independent replication. Further inquiry regarding the specific nature of the AI model can be made by relevant parties to the corresponding author.

\bibliographystyle{unsrt}  
\bibliography{references}

\textbf{Acknowledgements.} We acknowledge the steering committee for CROSS-TRACKS for access to the data. For data acquisition, modeling, and validation, we thank the following: Julian Guldborg Birkemose, Christian Bang, Per Dahl Rasmussen, Anne Olsvig Boilesen, Lars Mellergaard, and Jacob Høy Berthelsen. For help with the data extraction pipelines, we thank Mike Pedersen. We also thank the rest of the Enversion team for their support. This work was also supported by the Innovation Fund Denmark [case number 8053-00076B].

\textbf{Author contributions}. S.M.L, J.L, M.J.J., and B.T. initiated the project. B.T., S.M.L., K.M.L, J.L, and M.J.J. contributed to the overall experimental design. S.M.L and M.K. created the dataset. S.M.L., M.K., M.V.O., and M.S.L. contributed to the software engineering and K.M.L., B.T., and S.M.L. analyzed the results. S.M.L. made the first paper draft. All authors contributed significantly to revision of the first paper draft, and approval of the final version of the manuscript. 

\textbf{Competing interests.} S.M.L, M.K., and B.T. are employed at Enversion. The authors have no other competing interests to disclose.

\begin{table}[!h]
\captionsetup{singlelinecheck = false, justification=justified}
\renewcommand{\tablename}{Extended Data Table}
\caption{List of clinical parameters }
\begin{tabular}{lll}
\toprule
\multicolumn{3}{c}{\textbf{Laboratory parameters}} \\ \midrule
P(aB)-Hydrogen carbonate & P(aB)-Potassium & P-Albumin \\
P(vB)-Hydrogen carbonate & B-Hemoglobin & P-Creatinine \\
P(aB)-pO2 & B-Neutrophils & P-Bilirubin \\
P(vB)-pCO2 & B-Eythrocyte sedimentation rate & P-Prolactin \\
P(aB)-pCO2 & B-Platelets & P-Glucose \\
P(aB)-pH & B-Leukocytes & P-C-reactive protein (CRP) \\
P(vB)-pH & B-Hemoglobin & Hb(B)-Hemoglobin A1c \\
P(aB)-Lactate & P-Sodium & Glomerular filtration rate (eGFR) \\
P(aB)-Sodium & P-Potassium &  \\
P(aB)-Chloride & P-Lactate dehydrogenase (LDH) &  \\ \midrule
\multicolumn{3}{c}{\textbf{Vital sign parameters}} \\ \midrule
Systolic blood pressure & Respiratory Frequency & SpO2 (Pulsoxymetry) \\
Diastolic blood pressure & Pulse & Temperature \\ \bottomrule
\end{tabular}
\end{table}

\begin{table}[!h]
\captionsetup{singlelinecheck = false, justification=justified}
\renewcommand{\tablename}{Extended Data Table}
\caption{Patient population description }
\label{Extended Data Table 2 }
\begin{tabular}{ll}
\toprule
Unique patients, no. & 66,288 \\
Unique admissions, no. & 163,050 \\
Age, median years & 55.2 \\
Gender, male, \% of total admissions & 45.86 \\
Length of stays, median hours & 153.6 \\
Laboratory measurements, average per admission & 39 \\
Hospital mortality, \% of total admissions & 0.85 \\
Sepsis present, \% of total admissions & 2.44 \\
Acute Kidney Injury present, \% of total admissions & 0.75 \\
Acute Lung Injury present, \% of total admissions & 1.68 \\ 
\bottomrule
\end{tabular}
\end{table}

\begin{table}[!h]
\captionsetup{singlelinecheck = false, justification=justified}
\renewcommand{\tablename}{Extended Data Table}
\caption{The Area Under The Receiver Operating Characteristics (AUROC) curve}
\resizebox{\textwidth}{!}{%
\begin{tabular}{cccccccccccccc}
\toprule
\multicolumn{2}{c|}{\textbf{}} & \multicolumn{4}{c|}{\textbf{Sepsis}} & \multicolumn{4}{c|}{\textbf{Acute Kidney Injury}} & \multicolumn{4}{c}{\textbf{Acute Lunge Injury}} \\ \midrule
\textbf{\begin{tabular}[c]{@{}c@{}}Hours \\ before onset\end{tabular}} & \textbf{Fold} & \textbf{TCN} & \textbf{SOFA} & \textbf{MEWS} & \textbf{GB-Vital} & \textbf{TCN} & \textbf{SOFA} & \textbf{MEWS} & \textbf{GB-Vital} & \textbf{TCN} & \textbf{SOFA} & \textbf{MEWS} & \textbf{GB-Vital} \\ \midrule
0 & 1 & 0.92 & 0.83 & 0.8 & 0.84 & 0.88 & 0.75 & 0.70 & 0.76 & 0.90 & 0.80 & 0.85 & 0.65 \\
0 & 2 & 0.95 & 0.84 & 0.82 & 0.83 & 0.87 & 0.74 & 0.68 & 0.78 & 0.90 & 0.80 & 0.89 & 0.61 \\
0 & 3 & 0.92 & 0.83 & 0.82 & 0.85 & 0.88 & 0.74 & 0.72 & 0.78 & 0.93 & 0.77 & 0.85 & 0.63 \\
0 & 4 & 0.90 & 0.83 & 0.83 & 0.85 & 0.9 & 0.75 & 0.71 & 0.76 & 0.9 & 0.77 & 0.81 & 0.65 \\
0 & 5 & 0.93 & 0.83 & 0.84 & 0.83 & 0.86 & 0.73 & 0.70 & 0.76 & 0.89 & 0.82 & 0.84 & 0.63 \\
3 & 1 & 0.88 & 0.80 & 0.78 & 0.81 & 0.85 & 0.74 & 0.68 & 0.70 & 0.85 & 0.74 & 0.85 & 0.63 \\
3 & 2 & 0.85 & 0.80 & 0.79 & 0.81 & 0.84 & 0.73 & 0.68 & 0.72 & 0.88 & 0.74 & 0.78 & 0.62 \\
3 & 3 & 0.85 & 0.79 & 0.78 & 0.81 & 0.85 & 0.74 & 0.70 & 0.72 & 0.88 & 0.72 & 0.78 & 0.61 \\
3 & 4 & 0.89 & 0.79 & 0.78 & 0.80 & 0.83 & 0.75 & 0.68 & 0.71 & 0.85 & 0.73 & 0.81 & 0.55 \\
3 & 5 & 0.87 & 0.8 & 0.80 & 0.81 & 0.86 & 0.73 & 0.68 & 0.75 & 0.89 & 0.75 & 0.85 & 0.63 \\
6 & 1 & 0.86 & 0.78 & 0.78 & 0.78 & 0.84 & 0.74 & 0.70 & 0.70 & 0.83 & 0.73 & 0.81 & 0.61 \\
6 & 2 & 0.85 & 0.77 & 0.77 & 0.77 & 0.85 & 0.72 & 0.68 & 0.69 & 0.82 & 0.69 & 0.78 & 0.63 \\
6 & 3 & 0.86 & 0.79 & 0.77 & 0.78 & 0.85 & 0.74 & 0.70 & 0.69 & 0.84 & 0.72 & 0.78 & 0.60 \\
6 & 4 & 0.86 & 0.76 & 0.78 & 0.80 & 0.81 & 0.74 & 0.69 & 0.70 & 0.85 & 0.75 & 0.78 & 0.59 \\
6 & 5 & 0.83 & 0.79 & 0.77 & 0.80 & 0.83 & 0.73 & 0.67 & 0.68 & 0.88 & 0.70 & 0.71 & 0.63 \\
12 & 1 & 0.84 & 0.77 & 0.68 & 0.78 & 0.85 & 0.72 & 0.61 & 0.68 & 0.82 & 0.70 & 0.74 & 0.60 \\
12 & 2 & 0.83 & 0.76 & 0.71 & 0.77 & 0.85 & 0.71 & 0.69 & 0.68 & 0.86 & 0.76 & 0.75 & 0.59 \\
12 & 3 & 0.84 & 0.77 & 0.71 & 0.78 & 0.86 & 0.72 & 0.66 & 0.65 & 0.84 & 0.76 & 0.83 & 0.61 \\
12 & 4 & 0.85 & 0.77 & 0.72 & 0.79 & 0.85 & 0.72 & 0.67 & 0.68 & 0.85 & 0.72 & 0.77 & 0.61 \\
12 & 5 & 0.84 & 0.75 & 0.72 & 0.79 & 0.83 & 0.71 & 0.61 & 0.65 & 0.82 & 0.70 & 0.80 & 0.61 \\
24 & 1 & 0.81 & 0.72 & 0.71 & 0.79 & 0.8 & 0.66 & 0.62 & 0.68 & 0.85 & 0.67 & 0.70 & 0.60 \\
24 & 2 & 0.81 & 0.72 & 0.71 & 0.79 & 0.77 & 0.66 & 0.61 & 0.60 & 0.84 & 0.65 & 0.71 & 0.60 \\
24 & 3 & 0.80 & 0.72 & 0.67 & 0.77 & 0.77 & 0.62 & 0.60 & 0.65 & 0.82 & 0.70 & 0.74 & 0.59 \\
24 & 4 & 0.82 & 0.65 & 0.65 & 0.77 & 0.81 & 0.67 & 0.62 & 0.63 & 0.85 & 0.67 & 0.76 & 0.58 \\
24 & 5 & 0.77 & 0.71 & 0.61 & 0.76 & 0.79 & 0.62 & 0.62 & 0.60 & 0.83 & 0.67 & 0.75 & 0.58 \\ \bottomrule
\end{tabular}%
}
\small{TCN: Temporal Convolutional Network, SOFA: Sequential Organ Failure Assessment, MEWS: Early Warning Score, GB-Vital: Gradient Boosting model based on vital sign parameters.}
\end{table}

\begin{table}[t]
\captionsetup{singlelinecheck = false, justification=justified}
\renewcommand{\tablename}{Extended Data Table}
\caption{Area Under The Precision Recall (AUPRC) Curve }
\resizebox{\textwidth}{!}{%
\begin{tabular}{cccccccccccccc}
\toprule
\multicolumn{2}{c|}{\textbf{}} & \multicolumn{4}{c|}{\textbf{Sepsis}} & \multicolumn{4}{c|}{\textbf{Acute Kidney Injury}} & \multicolumn{4}{c}{\textbf{Acute Lunge Injury}} \\ \midrule
\textbf{\begin{tabular}[c]{@{}c@{}}Hours \\ before onset\end{tabular}} & \textbf{Fold} & \textbf{TCN} & \textbf{SOFA} & \textbf{MEWS} & \textbf{GB-Vital} & \textbf{TCN} & \textbf{SOFA} & \textbf{MEWS} & \textbf{GB-Vital} & \textbf{TCN} & \textbf{SOFA} & \textbf{MEWS} & \textbf{GB-Vital} \\ \midrule
0 & 1 & 0.92 & 0.83 & 0.8 & 0.84 & 0.88 & 0.75 & 0.70 & 0.76 & 0.90 & 0.80 & 0.85 & 0.65 \\
0 & 2 & 0.95 & 0.84 & 0.82 & 0.83 & 0.87 & 0.74 & 0.68 & 0.78 & 0.90 & 0.80 & 0.89 & 0.61 \\
0 & 3 & 0.92 & 0.83 & 0.82 & 0.85 & 0.88 & 0.74 & 0.72 & 0.78 & 0.93 & 0.77 & 0.85 & 0.63 \\
0 & 4 & 0.90 & 0.83 & 0.83 & 0.85 & 0.9 & 0.75 & 0.71 & 0.76 & 0.9 & 0.77 & 0.81 & 0.65 \\
0 & 5 & 0.93 & 0.83 & 0.84 & 0.83 & 0.86 & 0.73 & 0.70 & 0.76 & 0.89 & 0.82 & 0.84 & 0.63 \\
3 & 1 & 0.88 & 0.80 & 0.78 & 0.81 & 0.85 & 0.74 & 0.68 & 0.70 & 0.85 & 0.74 & 0.85 & 0.63 \\
3 & 2 & 0.85 & 0.80 & 0.79 & 0.81 & 0.84 & 0.73 & 0.68 & 0.72 & 0.88 & 0.74 & 0.78 & 0.62 \\
3 & 3 & 0.85 & 0.79 & 0.78 & 0.81 & 0.85 & 0.74 & 0.70 & 0.72 & 0.88 & 0.72 & 0.78 & 0.61 \\
3 & 4 & 0.89 & 0.79 & 0.78 & 0.80 & 0.83 & 0.75 & 0.68 & 0.71 & 0.85 & 0.73 & 0.81 & 0.55 \\
3 & 5 & 0.87 & 0.8 & 0.80 & 0.81 & 0.86 & 0.73 & 0.68 & 0.75 & 0.89 & 0.75 & 0.85 & 0.63 \\
6 & 1 & 0.86 & 0.78 & 0.78 & 0.78 & 0.84 & 0.74 & 0.70 & 0.70 & 0.83 & 0.73 & 0.81 & 0.61 \\
6 & 2 & 0.85 & 0.77 & 0.77 & 0.77 & 0.85 & 0.72 & 0.68 & 0.69 & 0.82 & 0.69 & 0.78 & 0.63 \\
6 & 3 & 0.86 & 0.79 & 0.77 & 0.78 & 0.85 & 0.74 & 0.70 & 0.69 & 0.84 & 0.72 & 0.78 & 0.60 \\
6 & 4 & 0.86 & 0.76 & 0.78 & 0.80 & 0.81 & 0.74 & 0.69 & 0.70 & 0.85 & 0.75 & 0.78 & 0.59 \\
6 & 5 & 0.83 & 0.79 & 0.77 & 0.80 & 0.83 & 0.73 & 0.67 & 0.68 & 0.88 & 0.70 & 0.71 & 0.63 \\
12 & 1 & 0.84 & 0.77 & 0.68 & 0.78 & 0.85 & 0.72 & 0.61 & 0.68 & 0.82 & 0.70 & 0.74 & 0.60 \\
12 & 2 & 0.83 & 0.76 & 0.71 & 0.77 & 0.85 & 0.71 & 0.69 & 0.68 & 0.86 & 0.76 & 0.75 & 0.59 \\
12 & 3 & 0.84 & 0.77 & 0.71 & 0.78 & 0.86 & 0.72 & 0.66 & 0.65 & 0.84 & 0.76 & 0.83 & 0.61 \\
12 & 4 & 0.85 & 0.77 & 0.72 & 0.79 & 0.85 & 0.72 & 0.67 & 0.68 & 0.85 & 0.72 & 0.77 & 0.61 \\
12 & 5 & 0.84 & 0.75 & 0.72 & 0.79 & 0.83 & 0.71 & 0.61 & 0.65 & 0.82 & 0.70 & 0.80 & 0.61 \\
24 & 1 & 0.81 & 0.72 & 0.71 & 0.79 & 0.8 & 0.66 & 0.62 & 0.68 & 0.85 & 0.67 & 0.70 & 0.60 \\
24 & 2 & 0.81 & 0.72 & 0.71 & 0.79 & 0.77 & 0.66 & 0.61 & 0.60 & 0.84 & 0.65 & 0.71 & 0.60 \\
24 & 3 & 0.80 & 0.72 & 0.67 & 0.77 & 0.77 & 0.62 & 0.60 & 0.65 & 0.82 & 0.70 & 0.74 & 0.59 \\
24 & 4 & 0.82 & 0.65 & 0.65 & 0.77 & 0.81 & 0.67 & 0.62 & 0.63 & 0.85 & 0.67 & 0.76 & 0.58 \\
24 & 5 & 0.77 & 0.71 & 0.61 & 0.76 & 0.79 & 0.62 & 0.62 & 0.60 & 0.83 & 0.67 & 0.75 & 0.58 \\ \bottomrule
\end{tabular}%
}
\small{TCN: Temporal Convolutional Network, SOFA: Sequential Organ Failure Assessment, MEWS: Early Warning Score, GB-Vital: Gradient Boosting model based on vital sign parameters.}
\end{table}

\end{document}